\documentclass[journal]{IEEEtran}

\usepackage{graphicx}

\usepackage{amsmath}
\usepackage{amssymb}
\usepackage{verbatim}
\usepackage{booktabs}

\usepackage{multirow}
\usepackage{algorithm}
\usepackage{algorithmic}
\usepackage{color}
\usepackage[dvipsnames]{xcolor}
\usepackage[pagebackref,breaklinks,colorlinks]{hyperref}
\usepackage[capitalize]{cleveref}
\usepackage{flushend}
\usepackage{bbm}
\usepackage{colortbl}
\usepackage{subfigure}
\crefname{section}{Sec.}{Secs.}
\Crefname{section}{Section}{Sections}
\Crefname{table}{Table}{Tables}
\crefname{table}{Tab.}{Tabs.}

\definecolor{person}{rgb}{0.75,0.5,0.5}
\definecolor{car}{rgb}{0.5,0.5,0.5}
\definecolor{motorbike}{rgb}{0.25,0.5,0.5}
\definecolor{dog}{rgb}{0.25,0,0.5}
\definecolor{sofa}{rgb}{0,0.75,0}
\definecolor{monitor}{rgb}{0,0.25,0.5}
\definecolor{border}{rgb}{0.825,0.825,0.7}
\definecolor{train}{rgb}{0.5,0.75,0}
\definecolor{table}{rgb}{0.75,0.5,0}
\usepackage[T1]{fontenc}
\newenvironment{myitemize}{\begin{list}{$\bullet$}
		{\setlength{\topsep}{1mm}
			\setlength{\itemsep}{0.25mm}
			\setlength{\parsep}{0.25mm}
			\setlength{\itemindent}{0mm}
			\setlength{\partopsep}{0mm}
			\setlength{\labelwidth}{15mm}
			\setlength{\leftmargin}{4mm}}}{\end{list}}

\ifCLASSINFOpdf
\else
\fi
\begin{document}
%
\title{Gradient-Semantic Compensation for Incremental Semantic Segmentation}
%
%
%

\author{Wei~Cong,
        Yang Cong,~\IEEEmembership{Senior Member,~IEEE,}
        Jiahua Dong,
        Gan Sun,~\IEEEmembership{Member,~IEEE,}
        Henghui Ding
\thanks{Wei Cong and Jiahua Dong are with the State Key Laboratory of Robotics, Shenyang Institute of Automation, Chinese Academy of Sciences, Shenyang 110016, China, also with the Institutes for Robotics and Intelligent Manufacturing, Chinese Academy of Sciences, Shenyang 110169, China, and also with the University of Chinese Academy of Sciences, Beijing 100049, China (email: congwei45@gmail.com; dongjiahua1995@gmail.com).}
\thanks{Yang Cong and Gan Sun are with the State Key Laboratory of Robotics, Shenyang Institute of Automation, Chinese Academy of Sciences, Shenyang 110016, China, and also with the Institutes for Robotics and Intelligent Manufacturing, Chinese Academy of Sciences, Shenyang 110169, China (email: congyang81@gmail.com; sungan1412@gmail.com).}
\thanks{Henghui Ding is with Nanyang Technological University (NTU), Singapore 639798 (e-mail: henghui.ding@gmail.com).}
\thanks{Corresponding author: Prof. Yang Cong.}}

%
%

\markboth{Journal of \LaTeX\ Class Files,~Vol.~14, No.~8, August~2015}%
{Shell \MakeLowercase{\textit{et al.}}: Bare Demo of IEEEtran.cls for IEEE Journals}
%



\maketitle

\begin{abstract}
Incremental semantic segmentation aims to continually learn the segmentation of new coming classes without accessing the training data of previously learned classes. However, most current methods fail to address catastrophic forgetting and background shift since they 1) treat all previous classes equally without considering  different forgetting paces caused by imbalanced gradient back-propagation; 2) lack strong semantic guidance between classes. To tackle the above challenges, in this paper, we propose a \textbf{\underline{G}}radient-\textbf{\underline{S}}emantic \textbf{\underline{C}}ompensation (\textbf{GSC}) model, which surmounts incremental semantic segmentation from both gradient and semantic perspectives. Specifically, to address catastrophic forgetting from the gradient aspect, we develop a step-aware gradient compensation that can balance forgetting paces of previously seen classes via re-weighting gradient back-propagation. Meanwhile, we propose a soft-sharp semantic relation distillation to distill consistent inter-class semantic relations via soft labels for alleviating catastrophic forgetting from the semantic aspect.  In addition, we develop a prototypical pseudo re-labeling that provides strong semantic guidance to mitigate background shift. It produces high-quality pseudo labels for old classes in the background by measuring distances between pixels and class-wise prototypes. Extensive experiments on three public datasets, \emph{i.e.,} Pascal VOC 2012, ADE20K, and Cityscapes, demonstrate the effectiveness of our proposed GSC model. 
\end{abstract}


%
\IEEEpeerreviewmaketitle

\section{Introduction}
\label{sec:intro}

\begin{figure}[t]
  \centering
  \includegraphics[width=1\linewidth]{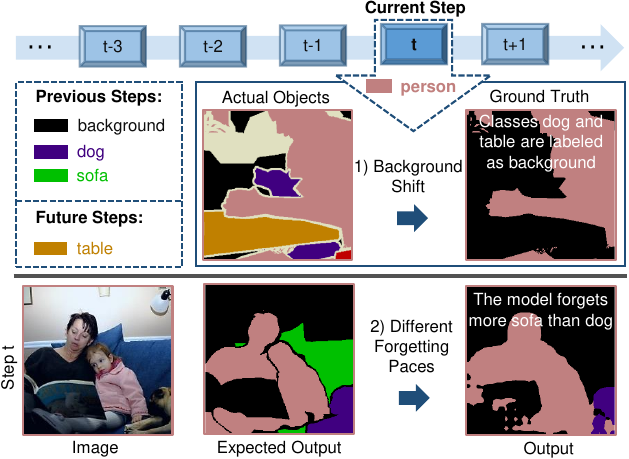}
   \caption{Illustration of challenges for incremental semantic segmentation (ISS). 1) Old classes (\emph{e.g.,} \textcolor{dog}{\texttt{dog}}) and future classes (\emph{e.g.,} \textcolor{table}{\texttt{table}}) are labeled as \texttt{background} in the current step $t$ where \textcolor{person}{\texttt{person}} is foreground, resulting in background shift. 2) The segmentation model continuously learns new classes without accessing to previous training samples, suffering from different forgetting paces of old classes (\emph{i.e.,} the model forgets more \textcolor{sofa}{\texttt{sofa}} than \textcolor{dog}{\texttt{dog}}). }
   \label{fig:motivation}
\end{figure}

\IEEEPARstart{S}{emantic} segmentation ~\cite{semantic_segmentation_survey} is one of the fundamental research fields in computer vision, which aims at classifying each pixel in an image and can be applicable to many fields, such as autonomous driving ~\cite{driving}. In general, all predefined classes in the standard fully-supervised segmentation setting are learned at once during the training phase~\cite{fullyconvolution,pyramid,atrousconvolution}, where there are no new classes during testing. However, new categories in real-world applications can be encountered on the go, and traditional segmentation methods cannot tackle new categories that have never observed in training. A trivial way is to finetune the segmentation model on the samples of new categories, which makes the existing methods brittle in an incremental learning (IL) setting ~\cite{EWC}. It is natural to seek effective incremental semantic segmentation (ISS) methods \cite{RCIL,STISS}, which can continually learn the model with the training samples of novel classes only. There have been some ISS works~\cite{MIB,PLOP,SDR,RCIL,STISS,UCD,CAF}, which mainly focus on two key challenges, catastrophic forgetting ~\cite{catastrophic_forgetting} and background shift ~\cite{MIB} (see \cref{fig:motivation}).

The first challenge, inherited from IL \cite{EWC,LWF}, is catastrophic forgetting~\cite{catastrophic_forgetting}, \emph{i.e.,} the knowledge of previously learned classes is abruptly lost, since the network weights are changed to meet the objectives of new classes. As shown in \cref{fig:motivation}, the segmentation model forgets a lot about the old classes \texttt{sofa} and \texttt{dog} when learning about the new class \texttt{person}. Most current methods mitigate catastrophic forgetting by 1) simply distilling the features or output probabilities of the old model to the new model~\cite{ILT, PLOP, RCIL}; 2) enforcing the latent space consistency via prototype matching and contrastive learning~\cite{SDR}. These methods start from constraining the consistency of old features by equally treating all old classes. However, the forgetting paces vary significantly since the gradient back-propagation of old classes is affected discriminatively by new classes. As shown in \cref{fig:motivation}, the segmentation model could totally forget the old class \texttt{sofa} but still remembers the old class \texttt{dog}. Thus, it is essential to balance the forgetting paces of old classes from \textit{\textbf{the gradient aspect}}. In this paper, we propose a step-aware gradient compensation to discriminatively treat old classes at different learning steps via re-weighting their gradient back-propagation of the last output layer. Precisely, the step-aware weight is calculated by the normalized gradient according to the average gradient of old classes at different incremental learning steps. Then we integrate the step-aware weight with the cross-entropy loss based on pseudo labels. It re-weights gradient back-propagation of old classes (\emph{i.e.,} increases the gradient back-propagation of forgotten classes and decreases the gradient back-propagation of remembered classes) to overcome catastrophic forgetting. Moreover, current methods~\cite{ILT,PLOP,RCIL,SDR} distill old knowledge without considering inter-class semantic similarity relations. To further alleviate catastrophic forgetting from \textit{\textbf{the semantic aspect}}, we propose a soft-sharp semantic relation distillation that takes inter-class semantic relations into consideration. Specifically, a soft semantic relation distillation loss is proposed to distill the inter-class semantic similarity relations via constructed soft labels, which ensures the inter-class semantic relations consistency across different incremental learning steps. Meanwhile, a sharp confidence loss is designed to make the output probabilities of the current segmentation model confident. 

The second challenge for ISS is the background shift~\cite{MIB}, \emph{i.e.,} all the pixels that do not belong to any current classes are assigned to the \texttt{background}. Thus, the background pixels contain three groups: the true background, the old object classes already learned, and the future classes not seen before. As illustrated in \cref{fig:motivation}, the old class \texttt{dog} and future class \texttt{table} are labeled as \texttt{background} in the current step $t$. Therefore, the semantic features of background pixels are changing over time, which risks exacerbating catastrophic forgetting. Some methods~\cite{MIB} convert the probabilities of the background class to the probabilities of previous classes or future classes, which fails to overcome background shift without labeling classes in the background. Other methods~\cite{PLOP, STISS} adopt the more reasonable pseudo-labeling technique to produce pseudo labels for old classes in the background. However, the pseudo labels are noisy without effective constraints, which will degrade model performance. In this paper, we propose a prototypical pseudo re-labeling, which leverages the distances between pixels and class-wise prototypes to remove misclassified pixels produced by the old model. It provides strong \textit{\textbf{semantic guidance}} to reduce background shift. 

In summary, the main contributions of this paper are:

\begin{myitemize}
\item We propose a \textbf{\underline{G}}radient-\textbf{\underline{S}}emantic \textbf{\underline{C}}ompensation (\textbf{GSC}) model to surmount incremental semantic segmentation. As we all know, this is an early attempt to consider gradient and semantic compensation for ISS.

\item To mitigate catastrophic forgetting, we design a step-aware gradient compensation to balance different forgetting paces of old classes from the gradient aspect. Moreover, a soft-sharp semantic relation distillation is proposed to keep the inter-class semantic relations consistent via constructed soft labels from the semantic aspect.

\item To tackle background shift, we propose a prototypical pseudo re-labeling to produce high-quality pseudo labels for old classes in the background, which targets at providing strong semantic guidance. 
\end{myitemize}

\section{Related Work}
\label{sec:related_work}

\textbf{Semantic Segmentation}
has achieved tremendous progress with the development of deep learning approaches~\cite{semantic_segmentation_survey,boundary}. Fully convolutional networks ~\cite{fullyconvolution} take any arbitrary size of input images and output segmentation maps, achieving remarkable results on several benchmarks~\cite{vocdataset,cityscapesdataset,adedataset}. Encoder-Decoder architectures~\cite{Deconvolution_Network,segnet,FBSNet} retain spatial information. 
The deconvolutional layer~\cite{Deconvolution_Network} and SegNet~\cite{segnet} are proposed to generate accurate maps and upsample the corresponding decoder, respectively. In addition, attention mechanisms~\cite{Attention_to_Scale,cross-channel}, multi-scale feature aggregation~\cite{ding2018context, multi-scale}, dilated convolution~\cite{atrousconvolution,deeplab}, and pyramid context aggregation~\cite{pyramid,Adaptive_Pyramid_Context} are designed to improve performance. Recently, Transformer ~\cite{transformer_segmentation} further promotes segmentation performance. However, these methods cannot tackle new encountered classes in real world. 

\textbf{Incremental Learning}
studies the problem of efficiently learning continuous tasks based on acquired knowledge without degrading performance on previous tasks~\cite{lifelong_machine,sg1,sg2}. According to the storage information in sequential learning process, we divide recent incremental learning algorithms into three categories: replay-based strategies, dynamic architecture-based strategies, and regularization-based strategies. Replay-based algorithms~\cite{icarl,GEM,AMSS,RBN,FCIL,uncertainty-d,emotion} learn a new task by merging the stored or generated old samples into the current training process.~\cite{PACKNET,DEN,DA,Audio} dynamically assign parameters for previous tasks to guarantee the stability of performance. Regularization-based strategies aim at constraining the model as it changes, either directly on network weights~\cite{EWC,MAS,SI,ONLINE-EWC}, output probabilities~\cite{LWF,FCIL,FISS}, or intermediary features\cite{CLASSIFIER}. We focus on the challenging incremental semantic segmentation.

\textbf{Incremental Semantic Segmentation}
aims at learning the tasks of semantic segmentation continually~\cite{ILT}.~\cite{ILT,PLOP,RCIL,dkd} propose to overcome forgetting by distilling intermediary features or output probabilities. Then, MiB~\cite{MIB} alleviates background shift via distilling the converted probabilities of the background. SDR~\cite{SDR} leverages prototype matching and contrastive learning to keep latent space consistent. PLOP~\cite{PLOP} and~\cite{STISS} develop pseudo-labeling approaches to boost performance. ~\cite{alife} presents a feature replay scheme and an adaptive regularizer to balance accuracy and efficiency. SSUL~\cite{SSUL} and MicroSeg~\cite{microseg} introduce additional pretrained models~\cite{detector,mask2former} to detect classes in the background. RCIL~\cite{RCIL} proposes a representation compensation module which is consisted of two dynamically evolved branches with one frozen and one trainable. EWF~\cite{EWF} fuses the model containing old knowledge and the model retaining new knowledge in a dynamic fusion manner, strengthening the memory of old classes in ever-changing distributions. However, these methods cannot tackle ISS challenges by treating old classes equally and lacking semantic guidance of old classes. In what follows, we introduce the design of our GSC model which re-weights gradient back-propagation and provides strong semantic guidance.

\begin{figure*}[htbp]
  \centering
  \includegraphics[width=1\linewidth]{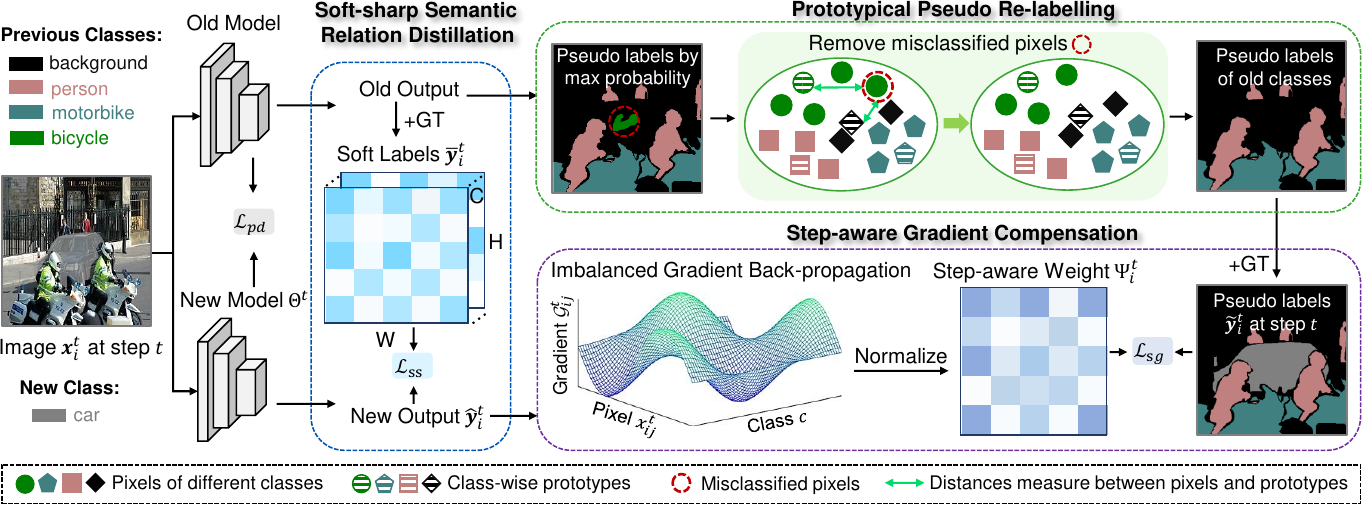}
   \caption{Overall framework of our \textbf{GSC} model. Given $\{\mathbf{x}^t_i,\mathbf{y}^t_i\}_{i=1}^{|\mathcal{D}^t|}\in \mathcal{D}^t$, the pseudo labels $\tilde{\mathbf{y}}^t_{i}$ are obtained by the strategy of prototypical pseudo re-labeling, and the soft labels $\bar{\mathbf{y}}^t_{i}$ are computed by combining the old output with the current one-hot ground truth. Then, using pseudo labels $\tilde{\mathbf{y}}^t_{i}$ and soft labels $\bar{\mathbf{y}}^t_{i}$ as targets, we update the parameters $\Theta^t$ for the new model with the proposed step-aware gradient compensation and soft-sharp semantic relation distillation.}
   \label{fig:framework}
\end{figure*}

\section{Preliminaries}
\label{preliminaries}

In the challenging incremental learning scenario, incremental semantic segmentation (ISS) considers learning a model continually at $t=0\dots T$ steps, at each of which there are only training samples of one dataset for the current step. Specifically, at step $t$, the model trains on the newly added dataset $\mathcal{D}^t=\{\mathbf{x}^t_i,\mathbf{y}^t_i\}_{i=1}^{|\mathcal{D}^t|}$, where $\mathbf{x}^t_i\in \mathbb{R}^{H\times W\times 3}$ denotes the image with the size of $W\times H$, $\mathbf{y}^t_i\in \mathcal{Y}^{H\times W}$ ($\mathcal{Y}$ represents label space) is the corresponding one-hot ground truth (GT) segmentation mask, and $|\mathcal{D}^t|$ represents the number of training data in the dataset $\mathcal{D}^t$. It is worth noting that the dataset $\mathcal{D}^t$ only contains the labels of current classes $\mathcal{C}^t$. $C^t=card(\mathcal{C}^t)-1$ is the cardinality of the current classes, excluding the background class. However, missing previous training data of $\{\mathcal{D}^0,\mathcal{D}^1\dots\mathcal{D}^{t-1}\}$ leads to catastrophic forgetting of previously learned knowledge at steps $0$ to $t-1$. Moreover, labeling both old classes $\mathcal{C}^{0:t-1}$ and future classes $\mathcal{C}^{t+1:T}$ as the background class $c^{bg}$ introduces background shift, further exacerbating catastrophic forgetting. Naturally, the model is typically a fully-convolutional network, consisting of a feature extractor $f^t(\cdot)$ and a classifier $g^t(\cdot)$. In addition, $y^t_{ij}$, $\hat y^t_{ij}$, and $\tilde y^t_{ij}$ represent the ground truth, the max output probability, and the pseudo label of the $j$-th pixel in the $i$-th image at step $t$, respectively. The goal of ISS is to predict all the object classes $\mathcal{C}^{0:t}$ seen over time. The loss function at the current step $t$ is defined as the following:
\begin{equation}
	\mathcal{L}_{ce}(\Theta^t)=-\frac{1}{|\mathcal{D}^t|}\frac{1}{WH}\sum_{i=1}^{|\mathcal{D}^t|}\sum_{j=1}^{WH}{\mathbf{y}}^t_{ij}\cdot \log\hat {\mathbf{y}}^{t}_{ij},
	\label{cross_entropy_loss}
\end{equation}
where $\mathcal{L}_{ce}(\cdot)$ denotes the standard cross-entropy loss used for supervised semantic segmentation, $\mathbf{y}^t_{ij}\in \mathbb{R}^{W,H,1+C^{t}}$ is the one-hot ground truth segmentation mask, and $\hat{\mathbf{y}}^t_{ij}=g^t\circ f^t(\mathbf x^t_{ij}) \in \mathbb{R}^{W,H,1+C^0+\cdots+C^t}$ represents the softmax output predicted segmentation mask of the $j$-th pixel in the $i$-th image at step $t$ over the current network parameters $\Theta ^t$.

\section{Method}
\label{sec:method}
Our proposed GSC model (see \cref{fig:framework}) overcomes catastrophic forgetting via a step-aware gradient compensation (\cref{sec:step_aware}) from the gradient aspect and a soft-sharp semantic relation distillation (\cref{sec:soft_sharp}) from the semantic aspect. Furthermore, it mitigates background shift under the semantic guidance of the prototypical pseudo re-labeling (\cref{sec:Prototypical_re}).

\subsection{Step-aware Gradient Compensation}
\label{sec:step_aware}
When continuously learning new classes, the imbalanced distribution between old classes in the current dataset $\mathcal{D}^t$ causes imbalanced gradient back-propagation of the last output layer in the current network $\Theta^t$~\cite{podnet,Mnemonics_Training}. This phenomenon leads to significantly different forgetting paces of old classes, which further worsens catastrophic forgetting for ISS. To overcome the above problem from the gradient aspect, as shown in \cref{fig:framework}, we develop a step-aware gradient compensation to compensate for different forgetting paces of old classes via re-weighting their imbalanced gradient back-propagation of the last output layer at different incremental steps. Specifically, the gradient measurement $\mathcal{G}^t_{ij}$ with respect to the $y^t_{ij}$-th neuron $\mathcal{N}^t_{y^t_{ij}}$ of the last output layer in the current network $\Theta^t$ for the pixel $(x^t_{ij},y^t_{ij})$ in the current dataset $\mathcal{D}^t$ is calculated by:
\begin{equation}
	\mathcal{G}^t_{ij}=\frac{\partial\mathcal{L}_{ce}(\hat{\mathbf y}^t_{ij}, \mathbf{y}^t_{ij})}{\partial\mathcal{N}^t_{y^t_{ij}}}=   (\hat{p}^t_{ij})_{y^t_{ij}}-1,                    
	\label{gradient_measurement}
\end{equation}
where $(\hat{p}^t_{ij})_{y^t_{ij}}$ is the $y^t_{ij}$-th sigmoid output probability of the $j$-th pixel in the $i$-th sample.
At the $t$-th step, we compute the average gradient $\mathcal{G}^{m}$ of old classes (excluding the background class $c^{bg}$) at each step $m (m<t)$. Since there is a background class $c^{bg}$ in each step, we compute the average gradient $\mathcal{G}^{bg}$ of the background class separately. Therefore, the objective function is formulated as the following:
\begin{equation}\small
	\mathcal{G}^{m}\!\!=\!\!\frac{1}{\sum_{i=1}^{|\mathcal{D}^t|}\sum_{j=1}^{WH}\mathbbm{1}{(\tilde{y}_{ij}^t}\in \mathcal{C}^m)} \sum_{i=1}^{|\mathcal{D}^t|}\sum_{j=1}^{WH} |\mathcal{G}_{ij}^t|\cdot\mathbbm{1}(\tilde{y}_{ij}^t\in \mathcal{C}^m),         
	\label{gradient_old}
\end{equation}
\begin{equation}\small
	\mathcal{G}^{bg}\!\!=\!\!\frac{1}{\sum_{i=1}^{|\mathcal{D}^t|}\sum_{j=1}^{WH}\mathbbm{1}{(\tilde{y}_{ij}^t}\in c^{bg})} \sum_{i=1}^{|\mathcal{D}^t|}\sum_{j=1}^{WH} |\mathcal{G}_{ij}^t|\cdot\mathbbm{1}(\tilde{y}_{ij}^t\in c^{bg}),
	\label{gradient_bg}
\end{equation}
where $\mathbbm{1}(\cdot)$ is the indicator function that outputs 1 if the condition is true, and 0 otherwise. $\tilde{y}_{ij}^t$ is obtained in \cref{pseudo_labels_prototype}.

\begin{figure}[t]
  \centering
  \includegraphics[width=0.7\linewidth]{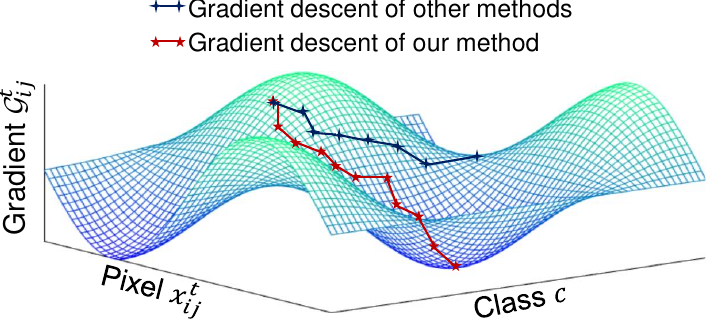}
   \caption{Illustration of the proposed step-aware gradient compensation, which makes the gradient descend better than other methods (\emph{e.g.,} PLOP and RCIL).}
   \label{fig:gradient}
\end{figure}

Then, according to the average gradient of old classes at different steps, the step-aware weight $\psi_{ij}^t$ is calculated by the normalized gradient of pixel $j$ in the current image $i$:
\begin{equation}
	\psi_{ij}^t=
	\left\{
	\begin{array}{lr}
		\frac{|\mathcal{G}_{ij}^t|}{\mathcal{G}^m}\;\text{if}\; \tilde{y}_{ij}^t \in \mathcal{C}^{m}\;\text{and}\; m<t\\
		\frac{|\mathcal{G}_{ij}^t|}{\mathcal{G}^{bg}}\;\text{if}\; \tilde{y}_{ij}^t \in c^{bg}\\
		\;\;1\;\;\;\;\text{otherwise}.
	\end{array}
	\right.
	\label{gradient_weight}
\end{equation}
We apply the constantly updated step-aware weight $\psi_{ij}^t$ to the cross-entropy loss based on one-hot pseudo labels $\tilde{\mathbf{y}}^t_{ij}$ in \cref{pseudo_labels_prototype}. And the different forgetting paces can be alleviated by re-weighting gradient back-propagation of old classes (see \cref{fig:gradient}). Thus, the step-aware gradient compensation loss is formulated as the following:
\begin{equation}
	\mathcal{L}_{sg}({\Theta}^t)= -\frac{1}{|\mathcal{D}^t|}\frac{1}{WH}\sum_{i=1}^{|\mathcal{D}^t|}\sum_{j=1}^{WH}\psi_{ij}^t(\tilde{\mathbf{y}}^t_{ij}\cdot \log\hat {\mathbf{y}}^{t}_{ij}).\\
	\label{gradient_entropy}
\end{equation}
The gradient $\mathcal{G}^t_{ij}$ for the pixel $j$ of an forgotten class is large, which causes $\psi_{ij}^t$ to be large in \cref{gradient_weight}. Then the large $\psi_{ij}^t$ will push the output probability of the pixel $j$ close to its pseudo label for reducing catastrophic forgetting in \cref{gradient_entropy}.

\subsection{Soft-sharp Semantic Relation Distillation}
\label{sec:soft_sharp}
We start with the learning mechanisms of human brains: when a human encounters an uncertain object, the human brain makes a preliminary judgment according to the inter-class similarities of the object with different classes (\emph{i.e.,} softness). Then the human brain gradually increases self-confidence and finally determines the category of the object (\emph{i.e.,} sharpness)~\cite{human}. Inspired by this, we propose a soft-sharp semantic relation distillation which is consisted of a soft semantic relation distillation loss and a sharp confidence loss to alleviate catastrophic forgetting by leveraging inter-class semantic relations from the semantic aspect.

\textbf{Soft Semantic Relation Distillation Loss:} During the learning of new classes, the model output probability reflects the underlying inter-class semantic relations between classes. To ensure the consistency of inter-class semantic relations, previous methods~\cite{ILT,LWF} only distill knowledge of old classes from the previous model to the current model. Different from previous methods that only ensure the semantic consistency of old classes, we propose a soft semantic relation distillation loss by distilling soft labels (\emph{i.e.,} semantic relations between old and new classes) to the output probability of the current model. Specifically, the soft label $\bar{\mathbf{y}}^t_{ij}$ is obtained by replacing the first $(1+C^0+\cdots+C^{t-1})$ dimensions of the one-hot ground truth segmentation mask $\mathbf{y}^t_{ij}$ with the sigmoid of $\hat{\mathbf{y}}^{t-1}_{ij}$ (the output probability of the old model), which effectively indicates the semantic relations between old and new classes by softly labeling old classes in the background. Therefore, the soft semantic relation distillation loss $\mathcal{L}_{sr}$ is:

\begin{equation}
	\mathcal{L}_{sr}(\Theta^t)= -\frac{1}{|\mathcal{D}^t|}\frac{1}{HW}\sum_{i=1}^{|\mathcal{D}^t|}\sum_{j=1}^{HW}\bar{\mathbf{y}}^t_{ij}\cdot \log\hat{\mathbf{y}}^{t}_{ij}.\\
	\label{soft_distillation}
\end{equation}
Despite the background difference between the current step and the last step, the current background contains a large number of foreground
pixels belonging to old classes. These pixels from old foreground classes have intrinsic inter-class semantic relations with new classes, and the soft label can well reflect such relations, which is effective to alleviate catastrophic forgetting via the distillation loss.

\textbf{Sharp Confidence Loss:}
By applying the distillation strategy, the model preserves inter-class semantic relations but leads to unconfident predictions, which is the dilemma of current methods~\cite{ILT,LWF} and does not conform to human learning habits (\emph{i.e.,} confidently identify the object class). Thus, we propose a sharp confidence loss to minimize the uncertainty of unconfident predictions and encourage the model predictions to be sharper, which is formulated as follows:

\begin{equation}
	\mathcal{L}_{sc}(\Theta^t)= -\frac{1}{|\mathcal{D}^t|}\frac{1}{HW}\sum_{i=1}^{|\mathcal{D}^t|}\sum_{j=1}^{HW}\hat{\mathbf{y}}^t_{ij}\cdot \log\hat{\mathbf{y}}^t_{ij}.\\
	\label{sharp_regularizer}
\end{equation}
Overall, the soft-sharp semantic relation distillation loss is:
\begin{equation}
	\mathcal{L}_{ss}(\Theta^t)= \lambda_1\mathcal{L}_{sr}(\Theta^t)+\lambda_2\mathcal{L}_{sc}(\Theta^t),\\
	\label{ss_distill}
\end{equation}
where $\lambda_1$ and $\lambda_2$ are the hyper-parameters to trade off the softness and sharpness of the model. Our soft-sharp semantic relation distillation can confidently predict the class while ensuring semantic relations between old and new classes.

\subsection{Prototypical Pseudo Re-labeling}
\label{sec:Prototypical_re}
Due to the changing of current classes $\mathcal{C}^t$ at each step, there is a background shift issue, \emph{i.e.,} background changes over steps while pixels labeled as background at step $t$ may belong to old classes, future classes, or the true background.
An effective way to mitigate the problem of background shift in ISS is to adopt the old model to predict pseudo labels~\cite{PLOP,STISS} for pixels in the background. However, there is a distribution shift of images among different steps. For the background, the closer to the distribution of old classes, the higher the confidence score by the old model. As a result, the distribution shift causes pseudo labels from the old model of previous methods~\cite{PLOP,STISS} to be noisy without effective constraints. Therefore, we rectify the pseudo labels of pixels for providing strong semantic guidance by estimating the class-wise likelihoods according to their relative feature distances to all class prototypes. In other words, if the feature $f^t(x^t_{ij})$ of image $\mathbf x^t_i$ at pixel $j$ is far from the prototype $\pmb{\eta}^c$ of class $c$, we down weight its probability of being classified into the $c$-th class. For example, in \cref{fig:framework}, before applying our strategy, the model misclassifies the pixels of \texttt{background} to ``\texttt{bicycle}''. By measuring the distances from the pixels to the prototypes of classes, we can remove the misclassified ``\texttt{bicycle}'' pixels. In specific, the class-wise correction weight $\zeta^t_{ij}(c)$ which measures distances between pixels $x^t_{ij}$ and prototypes of class $c$ is represented as:
\begin{equation}
  \zeta^t_{ij}(c)=\frac{\text{exp}(-||f^t(x^t_{ij})-\pmb{\eta}^{c}||/T)}{\sum_{c}\text{exp}(-||f^t(x^t_{ij})-\pmb{\eta}^{c}||/T)},
  \label{eq:weight_prototype}
\end{equation}
where the temperature $T=1$. Instead of directly learning new classes, we adopt the multi-scale pooling distillation loss $\mathcal{L}_{pd}(\Theta^t)$ ~\cite{PLOP} to keep the latent feature space consistent, leading to features of old classes extracted by the current model and the last model share the consistent feature space. Therefore, the prototypes are representative. There are two stages to obtain the prototype: 1) the coarse pseudo labels are the indexes of the max output probabilities by the old model over all pixels in the dataset $\mathcal{D}^t$; 2) according to the coarse pseudo labels, the features of the pixels predicted as $c$ in the background class are calculated to get the prototype $\pmb{\eta}^c$. The function of calculating prototype $\pmb{\eta}^c$ is:
\begin{equation}
	\pmb{\eta}^{c}=\frac{\sum f^{t-1}(x^t_{ij})*\mathbbm{1}(\hat {y}^{t-1}_{ij}(c)=1)}{\sum \mathbbm{1}(\hat {y}^{t-1}_{ij}(c)=1)},
	\label{prototype}
\end{equation}
where $\hat{\mathbf{y}}^t_{ij}(c)=1$ denotes the max output probability of the $j$-th pixel in the $i$-th image belonging to class $c$ at step $t$.

Based on the above, we propose an additional constraint related to prototypes for producing high-quality pseudo labels: the one-hot pseudo labels predicted by the old model, \emph{i.e.,} the max probability $\mathop{\arg\max}\limits_{c' \in \mathcal{C}^{0:t-1}} \hat {\mathbf{y}}^{t-1}_{ij}(c')$, are consistent with the rectified pseudo labels produced through the class-wise correction weight, \emph{i.e.,} $\mathop{\arg\max}\limits_{c' \in \mathcal{C}^{0:t-1}}\zeta^t_{ij}(c') \hat {\mathbf{y}}^{t-1}_{ij}(c')$.

The one-hot pseudo label $\tilde{\mathbf{y}}^t_i\in \mathbb{R}^{W,H,1+C^0+\cdots+C^{t}}$ at step $t$ is computed using the one-hot ground truth $\mathbf{y}^t_i\in \mathbb{R}^{W,H,1+C^{t}}$ and the softmax output probability $\hat{\mathbf{y}}^{t-1}_i\in \mathbb{R}^{W,H,1+C^0+\cdots+C^{t-1}}$ of the old model. Overall, the prototypical pseudo re-labeling strategy we propose is:
\begin{equation}\small
	\tilde{\mathbf{y}}^t_{ij}(c)\!=\!
	\left\{\!\!\!
	\begin{array}{lr}
		1\;\text{if}\; {\mathbf{y}}^t_{ij}(c^{bg})\!=\!0 \;\text{and}\; c\!=\!\mathop{\arg\max}\limits_{c' \in \mathcal{C}^t} {\mathbf{y}}^t_{ij}(c')\\
		1\;\text{if}\; {\mathbf{y}}^t_{ij}(c^{bg})\!=\!1 \;\text{and}\; c\!=\!\mathop{\arg\max}\limits_{c' \in \mathcal{C}^{0:t-1}} \hat {\mathbf{y}}^{t-1}_{ij}(c')
		\;\text{and}\; \\
		\;\;\; \mu\!<\!\tau^c\; \text{and}\; c\!=\!\mathop{\arg\max}\limits_{c' \in \mathcal{C}^{0:t-1}}\zeta^t_{ij}(c') \hat {\mathbf{y}}^{t-1}_{ij}(c')\\
		0\; \quad\quad\quad \text{otherwise},
	\end{array}
	\right.
	\label{pseudo_labels_prototype}
\end{equation}
where $\tilde{\mathbf{y}}^t_{ij}(c)=1$ denotes the pseudo label of the $j$-th pixel in the $i$-th image belonging to class $c$ at step $t$. $\mu$ represents the uncertainty of the pixel $j$. $\tau^c$ is the median entropy of class $c$, which is computed by the old segmentation model over all pixels of the dataset $\mathcal{D}^t$ predicted as class $c$.

It can be seen from \cref{pseudo_labels_prototype} that the prototypical pseudo re-labeling strategy at step $t$ can be divided into three parts: 1) all non-background pixels are labeled as the ground truth label; 2) if the old model is ``confident'' enough, the background pixels are labeled as its real semantic label; 3) if the old model is ``unconfident'', the corresponding background pixels are discarded. By performing the prototypical pseudo re-labeling strategy, we obtain more high-quality pseudo labels of the old classes, which provides strong semantic guidance to alleviate the background shift issue.

As mentioned before, we use the one-hot pseudo labels in \cref{pseudo_labels_prototype} as targets for the step-aware gradient compensation loss $\mathcal{L}_{sg}({\Theta}^t)$ in \cref{gradient_entropy}. 
In conclusion, the objective optimization function of our proposed GSC model is:
\begin{equation}
	\mathcal{L}(\Theta^t)= \mathcal{L}_{sg}(\Theta^t)+\mathcal{L}_{ss}(\Theta^t)+\lambda\mathcal{L}_{pd}(\Theta^t),\\
	\label{gsc}
\end{equation}
where $\lambda$ is the hyper-parameter for trading off losses. $\mathcal{L}_{pd}(\Theta^t)$ represents the multi-scale distillation loss~\cite{PLOP}.

\section{Experiments}
\label{sec:experiments}
\subsection{Experimental Setup}

\begin{table*}[htbp]
\caption{The mIoU(\%) of the last step on the Pascal VOC 2012 dataset for different class incremental segmentation scenarios. The \textbf{\color{purple}red} denotes the highest results and the \textbf{\color{blue}blue} denotes the second highest results. $*$ represents results from re-implementation.}
\setlength\tabcolsep{5pt}
\centering
\small
\scalebox{0.915}{
\begin{tabular}{ccc|c|cc|c|cc|c|cc|c|cc|c|cc|c}
\toprule

\multirow{3}{*}{{\textbf{Method}}}& \multicolumn{6}{c|}{\textbf{19-1 (2 steps)}}                                     & \multicolumn{6}{c|}{\textbf{10-10 (2 steps)}}                                     & \multicolumn{6}{c}{\textbf{15-1 (6 steps)}}                                     \\ \cmidrule{2-19}
          & \multicolumn{3}{c|}{\textbf{Disjoint}} & \multicolumn{3}{c|}{\textbf{Overlapped}} & \multicolumn{3}{c|}{\textbf{Disjoint}} & \multicolumn{3}{c|}{\textbf{Overlapped}} & \multicolumn{3}{c|}{\textbf{Disjoint}} & \multicolumn{3}{c}{\textbf{Overlapped}} \\ \cmidrule{2-19}
 & 0-19        & 20         & all        & 0-19         & 20          & all         & 0-10       & 11-20       & all        & 0-10        & 11-20        & all        & 0-15        & 16-20       & all       & 0-15        & 16-20        & all        \\ 
\midrule
\multicolumn{1}{c|}{FT}              & \, 5.8         & 12.3       & \, 6.2        & \, 6.8          & 12.9        & \, 7.1        & \, 7.7        & 60.8        & 33.0       & \, 7.8         & 58.9         & 32.1        & \, 0.2         & \, 1.8         & \, 0.6       & \, 0.2         & \, 1.8          & \, 0.6        \\ 
\multicolumn{1}{c|}{Joint}           & 77.4        & 78.0       & 77.4       & 77.4         & 78.0        & 77.4       & 78.6       & 76.0        & 77.4       & 78.6        & 76.0         & 77.4       & 79.1        & 72.6        & 77.4      & 79.1        & 72.6         & 77.4       \\ 
\multicolumn{1}{c|}{EWC~\cite{EWC}}             & 23.2        & 16.0       & 22.9       & 26.9         & 14.0        & 26.3       & \, 7.6        & 66.5        & 35.6      & \, 7.5        & 64.2         & 34.5       & \, 0.3         & \, 4.3         & \, 1.3       & \, 0.3         & \, 4.3          & \, 1.3        \\ 
\multicolumn{1}{c|}{LWF~\cite{LWF}}             & 53.0        & \, 9.1        & 50.8       & 51.2         & \, 8.5         & 49.1       & 63.1       & 61.1        & 62.2       & 70.7        & 63.4         & 67.2       & \, 0.8         & \, 3.6         & \, 1.5       & \, 1.0         & \, 3.9          & \, 1.8        \\ 
\multicolumn{1}{c|}{LWF-MC~\cite{icarl}}          & 63.0        & 13.2       & 60.5       & 64.4         & 13.3        & 61.9       & 52.4       & 42.5        & 47.7       & 53.9        & 43.0         & 48.7          & \, 4.5         & \, 7.0         & \, 5.2       & \, 6.4         & \, 8.4          & \, 6.9          \\ 
\multicolumn{1}{c|}{ILT~\cite{ILT}}             & 69.1        & 16.4       & 66.4       & 67.1         & 12.3        & 64.4       & 67.7       & 61.3        & \textbf{\color{blue}64.7}       & 70.3        & 61.9         & 66.3       & \, 3.7         & \, 5.7         & \, 4.2       & \, 4.9         & \, 7.8          & \, 5.7        \\ 
\multicolumn{1}{c|}{MiB~\cite{MIB}}             & 69.6        & 25.6       & 67.4       & 70.2         & 22.1        & 67.8       & 66.9       & 57.5        & 62.4       & 70.4        & 63.7         & 67.2       & 46.2        & 12.6        & 37.9      & 35.1        & 13.5         & 29.7       \\ 
\multicolumn{1}{c|}{SDR~\cite{SDR}}             & 69.9        & 37.3       & 68.4       & 69.1         & 32.6        & 67.4       & 67.5       & 57.9        & 62.9       & 70.5        & 63.9         & 67.4       & 59.2        & 12.9        & 48.1      & 44.7        & 21.8         & 39.2       \\ 
\multicolumn{1}{c|}{PLOP~\cite{PLOP}}            & 75.7        & 29.3       & \textbf{\color{blue}73.5}       & 75.4         & 37.4        & \textbf{\color{blue}73.5}       & 61.8       & 53.1        & 57.5       & 65.0        & 58.8         & 61.9       & 57.9        & 13.7        & 46.5      & 65.1        & 21.1         & 54.6       \\ 
\multicolumn{1}{c|}{RCIL$^*$~\cite{RCIL}}        & 73.8        & 27.1       & 71.6       & 74.9         & 32.3        & 72.8       & 65.1       & 42.1        & 54.1       & 74.2        & 60.7         & \textbf{\color{blue}67.8}       & 62.2        & 18.1        & 51.7      & 68.5        & 18.7         & 56.6       \\
\multicolumn{1}{c|}{RCIL~\cite{RCIL}}        & -        & -       & -      & -         & -        & -       & -       & -        & -       & -        & -         & -       & 66.1        & 18.2        & \textbf{\color{blue}54.7}      & 70.6        & 23.7         & \textbf{\color{blue}59.4}       \\
\midrule

\multicolumn{1}{c|}{\textbf{GSC} (ours)}   & 75.9        & 31.0       & \textbf{\color{purple}74.0}       & 76.9         & 42.7        & \textbf{\color{purple}75.3}       & 68.6           & 63.4            & \textbf{\color{purple}66.1}           & 76.8            & 63.2             & \textbf{\color{purple}70.3}       & 67.2        & 19.2        & \textbf{\color{purple}55.8}      & 72.1        & 24.4         & \textbf{\color{purple}60.8}       \\ 
\bottomrule

\end{tabular}
}
\label{tab:voc_base}
\end{table*}

\begin{figure*}[ht]
  \centering
  \includegraphics[width=1
\linewidth]{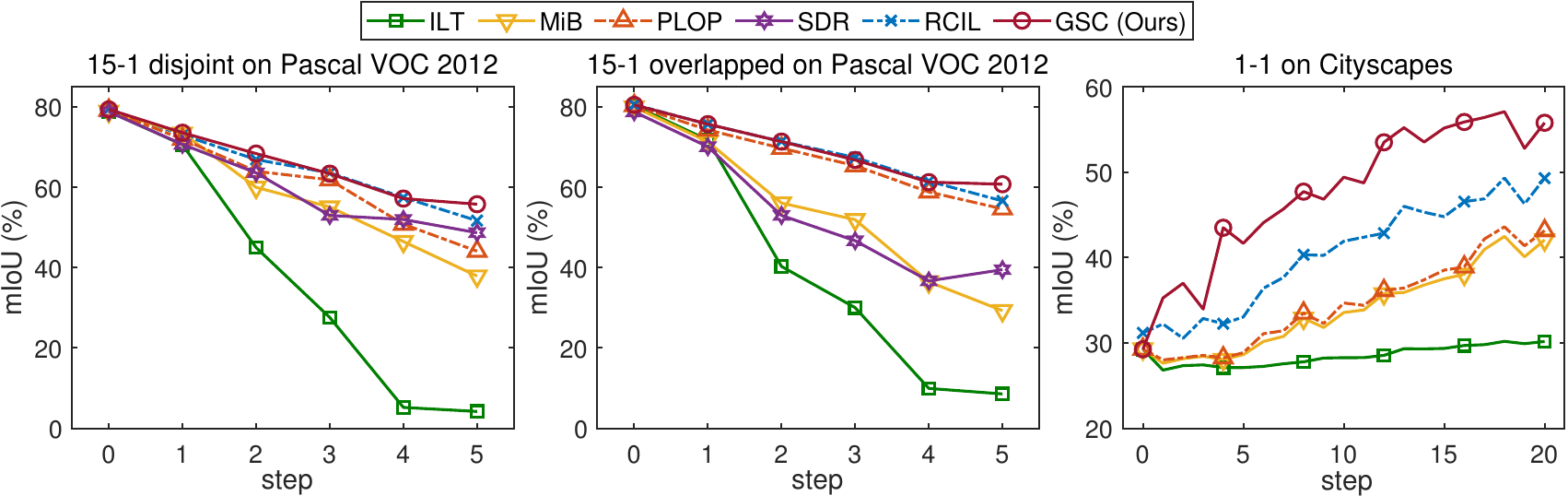} 
   \caption{The mIoU (\%) at each step in three incremental semantic segmentation scenarios. Incremental class segmentation: 15-1 disjoint on Pascal VOC 2012 (left), 15-1 overlapped on Pascal VOC 2012 (middle). Incremental domain segmentation: 1-1 on Cityscapes (right).}
   \label{fig:he}
\end{figure*}

\begin{table*}[t]
\caption{The mIoU(\%) of the last step on Pascal VOC 2012 for more incremental class learning scenarios. \textbf{\color{purple}Red}: the highest results. \textbf{\color{blue}Blue}: the second highest results. $*$: by re-implementation.}
\setlength\tabcolsep{9.7pt}
\centering
\small
\scalebox{0.955}{
\begin{tabular}{ccc|c|cc|c|cc|c|cc|c}
\toprule
\multirow{3}{*}{{\textbf{Method}}}& \multicolumn{6}{c|}{\textbf{15-5 (2 steps)}}                                              & \multicolumn{6}{c}{\textbf{10-1 (11 steps)}}                                               \\ \cmidrule{2-13}
                  & \multicolumn{3}{c|}{\textbf{Disjoint}} & \multicolumn{3}{c|}{\textbf{Overlapped}} & \multicolumn{3}{c|}{\textbf{Disjoint}} & \multicolumn{3}{c}{\textbf{Overlapped}} \\ \cmidrule{2-13}
& 0-15        & 16-20       & all       & 0-15        & 16-20        & all        & 0-10       & 11-20       & all        & 0-10        & 11-20        & all        \\ \midrule
\multicolumn{1}{c|}{FT}       & \ \ 1.1         & 33.6        & \ \ 9.2       & \ \ 2.1         & 33.1         & \ \ 9.8       & \ \ 6.3        & \ \ 1.1         & \ \ 3.8        & \ \ 6.4         & \ \ 1.2          & \ \ 3.9        \\ 
\multicolumn{1}{c|}{Joint}    & 79.1        & 72.6        & 77.4      & 79.1        & 72.6         & 77.4       & 78.6       & 76.0        & 77.4       & 78.6        & 76.0         & 77.4       \\ 
\multicolumn{1}{c|}{EWC~\cite{EWC}}     & 26.7        & 37.7        & 29.4      & 24.3        & 35.5         & 27.1        & \ \ 6.7        & \ \ 4.4         & \ \ 5.6        & \ \ 6.6        & \ \ 4.9         & \ \ 5.8       \\
\multicolumn{1}{c|}{LWF~\cite{LWF}}      & 58.4        & 37.4        & 53.1      & 58.9        & 36.6         & 53.3        & \ \ 7.2        & \ \ 1.2         & \ \ 4.3        & \ \ 8.0        & \ \ 2.0         & \ \ 4.8       \\ 
\multicolumn{1}{c|}{LWF-MC~\cite{icarl}}   & 67.2        & 41.2        & 60.7      & 58.1        & 35.0         & 52.3       & \ \ 6.9        & \ \ 1.7         & \ \ 4.4        & 11.2        & \ \ 2.5          & \ \ 7.1        \\ 
\multicolumn{1}{c|}{ILT~\cite{ILT}}      & 63.2        & 39.5        & 57.3      & 66.3        & 40.6         & 59.9       & \ \ 7.3       & \ \ 3.2         & \ \ 5.4        & \ \ 7.2        & \ \ 3.7          & \ \ 5.5        \\ 
\multicolumn{1}{c|}{MiB~\cite{MIB}}      & 71.8        & 43.3        & 64.7      & 75.5        & 49.4         & 69.0       & \ \ 9.5       & \ \ 4.1         & \ \ 6.9       & 20.0        & 20.1         & 20.1       \\ 
\multicolumn{1}{c|}{SDR~\cite{SDR}} & 73.5        & 47.3        & 67.2      & 75.4        & 52.6         & 69.9       & 17.3       & 11.0        & 14.3       & 32.4        & 17.1         & 25.1       \\
\multicolumn{1}{c|}{PLOP~\cite{PLOP}}     & 71.0        & 42.8        & 64.3      & 75.7        & 51.7         & 70.1       & \ \ 9.7       & \ \ 7.0        & \ \ 8.4       & 44.0        & 15.5         & 30.5       \\ 
\multicolumn{1}{c|}{RCIL$^*$~\cite{RCIL}} & 74.2        & 41.3        & 66.4      & 76.3        & 49.1         & 69.9       & 33.0       & \ \ 1.6        & 18.1       & 47.8        & 15.8         & 32.5       \\
\multicolumn{1}{c|}{RCIL~\cite{RCIL}} & 75.0        & 42.8        & \textbf{\color{blue}67.3}      & 78.8        & 52.0         & \textbf{\color{blue}72.4}      & 30.6       & \ \ 4.7        & \textbf{\color{blue}18.2}       & 55.4        & 15.1         & \textbf{\color{blue}34.3}       \\ \midrule
\multicolumn{1}{c|}{\textbf{GSC} (ours)}     & 74.4        & 45.8        & \textbf{\color{purple}67.6}      & 78.3        & 54.2         & \textbf{\color{purple}72.6}           & 34.2           & \ \ 6.1            & \textbf{\color{purple}20.8}           & 50.6            & 17.3             & \textbf{\color{purple}34.7}           \\ 
\bottomrule
\end{tabular}}

\label{tab:voc_increment}
\end{table*}

\begin{table*}[htbp]
\caption{The mIoU(\%) of the last step on the ADE20K dataset for overlapped incremental class segmentation scenarios, \emph{i.e.,} 100-50 (2 steps), 100-10 (6 steps) and 50-50 (3 steps). The \textbf{\color{purple}red} denotes the highest results and the \textbf{\color{blue}blue} denotes the second highest results.}
\setlength\tabcolsep{5.1pt}
\centering
\small
\scalebox{0.94}{
\begin{tabular}{ccc|c|cccccc|c|ccc|c}
\toprule
  \multirow{2}{*}{\textbf{Method}}     & \multicolumn{3}{c|}{\textbf{100-50 (2 steps)}} & \multicolumn{7}{c|}{\textbf{100-10 (6 steps)}}                           & \multicolumn{4}{c}{\textbf{50-50 (3 steps)}} \\ \cmidrule{2-15}
 & 1-100      & 101-150      & all      & 1-100 & 101-110 & 111-120 & 121-130 & 131-140 & 141-150 & all  & 1-50   & 51-100  & 101-150  & all   \\ \midrule
\multicolumn{1}{c|}{Joint}                   & 44.3       & 28.2         & 38.9     & 44.3  & 26.1    & 42.8    & 26.7    & 28.1    & 17.3    & 38.9 & 51.1   & 38.3    & 28.2     & 38.9  \\ 
\multicolumn{1}{c|}{ILT~\cite{ILT}}                     & 18.3       & 14.8         & 17.0     & \, 0.1   & \, 0.0     & \, 0.1     & \, 0.9     & \, 4.1     & \, 9.3     & \, 1.1  & 13.6   & 12.3    & \, 0.0      & \, 9.7   \\ 
\multicolumn{1}{c|}{MiB~\cite{MIB}}                     & 40.7       & 17.7         & 32.8     & 38.3  & 12.6    & 10.6    & \, 8.7     & \, 9.5     & 15.1    & 29.2 & 45.3   & 26.1    & 17.1     & 29.3  \\ 
\multicolumn{1}{c|}{PLOP~\cite{PLOP}}                    & 41.9       & 14.9         & 32.9     & 40.6  & 15.2    & 16.9    & 18.7    & 11.9    & \, 7.9     & 31.6 & 48.6   & 30.0    & 13.1     & 30.4  \\ 
\multicolumn{1}{c|}{RCIL~\cite{RCIL}}                    & 42.3       & 18.8         & \textbf{\color{blue}34.5}     & 39.3  & 14.6    & 26.3    & 23.2    & 12.1    & 11.8    & \textbf{\color{blue}32.1} & 48.3   & 31.3    & 18.7     & \textbf{\color{blue}32.5}  \\ \midrule
\multicolumn{1}{c|}{\textbf{GSC} (ours)}                   & 42.4           &19.2              & \textbf{\color{purple}34.8}         &40.8       &14.3         &24.6         &22.2        & 15.2        & 11.7        &\textbf{\color{purple}32.6}     &46.2        &30.2        &22.2          &\textbf{\color{purple}33.0}       \\ 
\bottomrule
\end{tabular}}

\label{tab:ade}
\end{table*}

\begin{figure*}[ht]
\centering
\hspace{-3mm}
\subfigure[Image]{
\label{fig:subfig:a}
\includegraphics[scale=0.93]{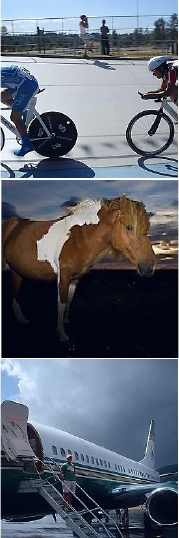}}
\hspace{-2.6mm}
\subfigure[MiB~\cite{MIB}]{
\label{fig:subfig:c}
\includegraphics[scale=0.93]{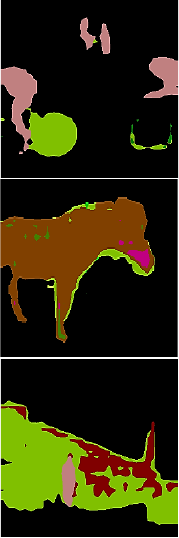}}
\hspace{-2.6mm}
\subfigure[PLOP~\cite{PLOP}]{
\label{fig:subfig:d}
\includegraphics[scale=0.93]{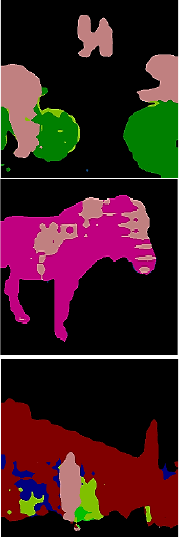}}
\hspace{-2.6mm}
\subfigure[RCIL~\cite{RCIL}]{
\label{fig:subfig:e}
\includegraphics[scale=0.93]{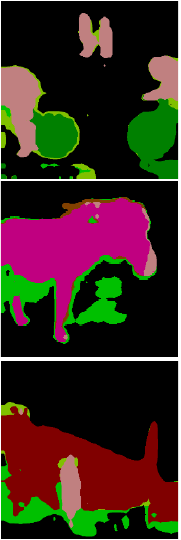}}
\hspace{-2.6mm}
\subfigure[\textbf{GSC} (Ours)]{
\label{fig:subfig:f}
\includegraphics[scale=0.93]{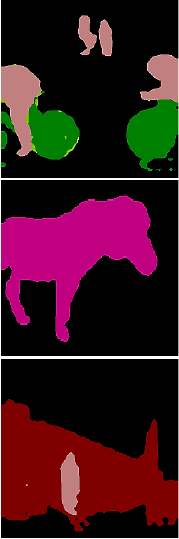}}
\hspace{-2.6mm}
\subfigure[GT]{
\label{fig:subfig:b}
\includegraphics[scale=0.93]{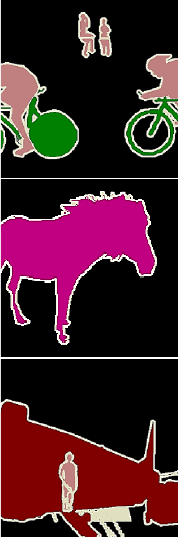}} 
\caption{The qualitative comparison between state-of-the-art methods. All the prediction results are from the last step of 15-1 overlapped scenario on Pascal VOC 2012 dataset. The visualization results demonstrate the effectiveness and superiority of the proposed GSC.}
\label{fig:vis}
\end{figure*}

\textbf{Datasets:} We evaluate our proposed GSC model on Pascal VOC 2012 ~\cite{vocdataset} (20 classes plus the background class), ADE20K ~\cite{adedataset} (150 classes) and Cityscapes ~\cite{cityscapesdataset} (19 classes from 21 different cities) datasets. Pascal VOC 2012 contains 10582 training images and 1449 testing images. ADE20K consists of 20210 and 2000 images for training and testing, respectively. There are 2975 images for training and 500 images for testing in Cityscapes. Specifically, we resize the images to 512×512 with a random resize crop and apply an additional random horizontal flip augmentation for all three datasets at the training time. Moreover, the images are resized to 512×512 with a center crop when testing.

\textbf{Competing Algorithms:} We compare our proposed GSC model with several classical incremental learning methods. \textbf{EWC}~\cite{EWC} constrains the update of essential model parameters when learning continuous data. \textbf{LWF}~\cite{LWF} transfers knowledge by distilling knowledge from the previous model to the new model. \textbf{LWF-MC}~\cite{icarl} replays the closest old samples to each class's feature mean and distills the old model's prediction to that of the current model. Moreover, we conduct experiments on several state-of-the-art incremental semantic segmentation (ISS) methods. \textbf{ILT}~\cite{ILT} distills the previous model's intermediary features and output probabilities to the current model. \textbf{MiB}~\cite{MIB} converts the likelihood of the background to the probability sum of old or future classes for handling background shift. \textbf{SDR}~\cite{SDR} regularizes the latent feature space by prototype matching and contrastive learning. \textbf{PLOP}~\cite{PLOP} distills multi-scale features and labels the background based on entropy. 
\textbf{RCIL}~\cite{RCIL} decomposes the representation learning into the old and new knowledge, and designs a pooled cube knowledge distillation strategy. Besides, two baselines are also considered. \textbf{Fine-tuning (FT)} is sequentially trained on the newly coming training data in the standard way, which can be seen as a lower bound. \textbf{Joint} is always trained using all training data seen so far, which can be seen as an upper bound.

\textbf{Implementation Details:}
Following ~\cite{MIB,PLOP}, we use Deeplab-v3~\cite{deeplab} with a ResNet-101 backbone~\cite{resnet} pre-trained on ImageNet~\cite{imagenet} for all experiments. The experiments are implemented in Pytorch and conducted on two Nvidia RTX 8000 GPUs. Each step contains 30 epochs for Pascal VOC 2012 and 60 epochs for ADE20K and Cityscapes, with a batch size of 24 distributed on two GPUs. The learning rate of the first step in all experiments is $1e-2$, and that of the following steps are $1e-3$ for Pascal VOC 2012 and $4e-3$ for ADE20K and Cityscapes. We reduce the learning rate exponentially with a decay rate of $9e-1$. We adopt SGD optimizer with $9e-1$ Nesterov momentum. $\lambda$ is set to 0.01, 0.001 and 0.0001 for Pascal VOC 2012, ADE20K and Cityscapes, respectively. In addition, the hyper-parameters $\lambda_1$ and $\lambda_2$ are set to 0.3 and 0.1, respectively. The first step is common to all methods for each setting, thus we reuse the weights trained in this step. In the phase of inference, no task id is provided, which is more realistic than some continual learning methods~\cite{EWC,SI}. 

\textbf{Evaluation Protocols:} 
Following~\cite{MIB} and~\cite{PLOP}, we consider incremental class segmentation and incremental domain segmentation. There are two settings for incremental class segmentation, \textit{disjoint} and \textit{overlapped}. If the pixels in the background class $c^{bg}$ at step $t$ belong to the classes $\mathcal{C}^{0:t-1}$ (\emph{i.e.,} old classes) and the true background class $c^{bg}$, it is a \textit{disjoint} setting. If the pixels in the background class $c^{bg}$ belong to classes $\mathcal{C}^{0:t-1}\cup \mathcal{C}^{t+1:T}$ (\emph{i.e.,} old and future classes) and the true background class $c^{bg}$, it is an \textit{overlapped} setting. The testing classes are labeled for all seen classes to evaluate the current model's incremental learning performance. On Pascal VOC 2012 dataset~\cite{vocdataset}, 19-1 (19 then 1 class), 15-5 (15 then 5 classes), 15-1 (15 classes followed by 1 class five times), 10-10 (10 then 10 classes) and 10-1 (10 classes followed by 1 class ten times) scenarios are conducted to measure the model performance. Among them, 10-1 is the most challenging scenario, which contains 11 steps. On ADE20K dataset~\cite{adedataset}, we conduct experiments on three scenarios: 100-50, 50-50 and 100-10. Incremental domain segmentation~\cite{incremental_domain,general_domain} aims at dealing with domain shifts instead of learning new classes. It studies the problem of different input spaces but the same output space with the same classes. Moreover, we conduct experiments of incremental domain segmentation on Cityscapes~\cite{cityscapesdataset}. Similarly, we apply three scenarios: 11-5, 11-1 and 1-1 with 3, 11 and 21 steps, respectively.

\textbf{Metrics:} We use mean Intersection over Union (mIoU) to measure the segmentation performance. Three results are computed after the last step $T$: 1) we compute mIoU for the initial classes $\mathcal{C}_0$, which reflects the competence to overcome catastrophic forgetting; 2) we compute mIoU for the incremental classes $\mathcal{C}_{1:T}$ for measuring the capacity of adapting to new classes; 3) we compute mIoU for all classes $\mathcal{C}_{0:T}$, which measures the overall continual learning ability.

\subsection{Incremental Class Segmentation}
\begin{figure*}[htbp]
  \centering
  \includegraphics[width=1\linewidth]{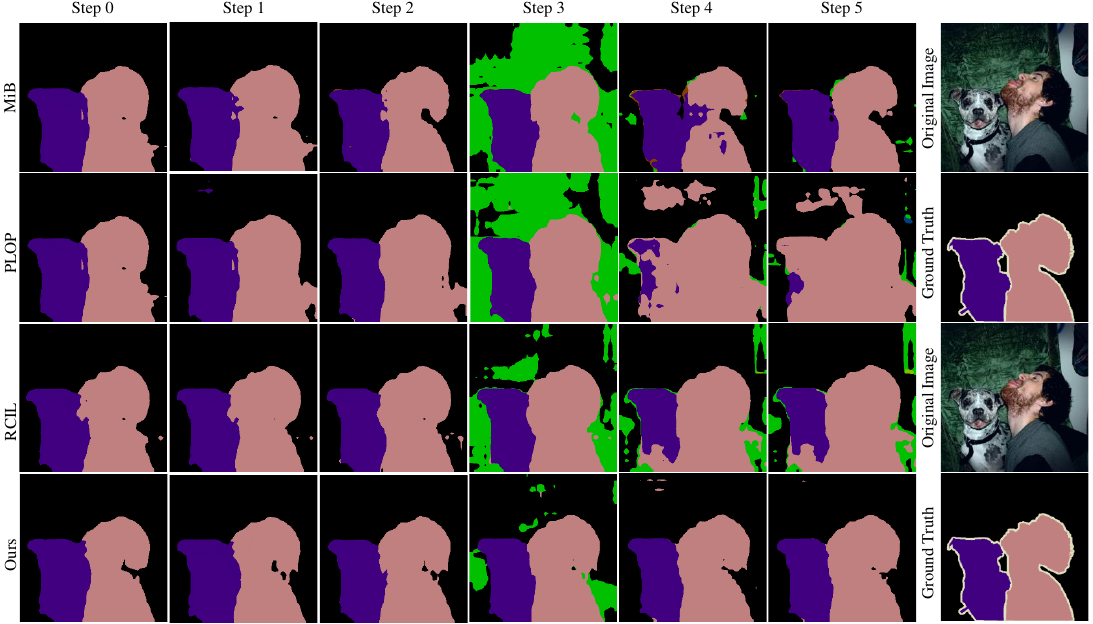}
   \caption{Visualization comparison between different methods across steps in the 15-1 overlapped scenario on the Pascal VOC 2012 dataset. MiB, Plop and RCIL (rows 1, 2 and 3) quickly forget the initial 15 classes (\textcolor{person}{\texttt{person}} and \textcolor{dog}{\texttt{dog}}) and is biased towards new classes (\textcolor{sofa}{\texttt{sofa}}). The proposed GSC model, however, barely suffers from the problem of catastrophic forgetting and background shift(rows 4).}
   \label{fig:step_vis}
\end{figure*}

\textbf{Pascal VOC 2012:} We perform experiments under \textit{disjoint} and \textit{overlapped} settings in terms of Pascal VOC 2012 dataset ~\cite{vocdataset}. We reproduce RCIL~\cite{RCIL} since it does not experiment in some scenarios. SSUL~\cite{SSUL} and MicroSeg~\cite{microseg} introduce pretrained detector~\cite{detector} and Mask2Former~\cite{mask2former} that have already seen classes in the background, which is not a fair comparison with our GSC model. \cref{tab:voc_base} shows results on 19-1, 10-10 and 15-1. We have the following observations: 
1) Our method excels in different class incremental semantic segmentation scenarios, \emph{i.e.,} \textit{disjoint} and \textit{overlapped} settings. Specifically, our method outperforms the state-of-the-art methods by 2.5\% in terms of mIoU in the 10-10 \textit{overlapped} scenario. 2) Our method has a considerable performance improvement on old classes, \emph{e.g.,} 2.6\% mIoU improvement in the 10-10 \textit{overlapped} scenario. This is benefited from the proposed step-aware gradient compensation, soft-sharp semantic relation distillation and prototypical pseudo re-labeling, which overcomes catastrophic forgetting and background shift from gradient and semantic perspectives.  3) The performance of our method on new classes outperforms other methods by 5.3\% mIoU and 1.0\% mIoU in the 19-1 \textit{overlapped} and 15-1 \textit{disjoint} settings, respectively. This indicates that our method makes room for new classes. We display the performance of each step for different methods in \cref{fig:he} (left and middle), which demonstrates the effectiveness of our GSC model at each step in ISS. Intuitively, We present the visualization results of the last step in the 15-1 \textit{overlapped} scenario in \cref{fig:vis}. Ours achieves less forgetting on old classes, illustrating that the proposed GSC model reduces the old-class forgetting and new-class overfitting. \cref{fig:step_vis} shows the predictions of the 15-1 \textit{overlapped} scenario on the Pascal VOC 2012 dataset for each step. At first, all methods output equivalent predictions, since there is no difference between the first step of different methods. However, other methods quickly forget previous classes and become biased toward new classes. Compared with them, the predictions of our proposed GSC model are much more stable in old classes while learning new classes. This is thanks to the proposed step-aware gradient compensation, soft-sharp semantic relation distillation and prototypical pseudo re-labeling strategies. 
Experiments on more challenging scenarios (\emph{i.e.,} 15-5 and 10-1) on Pascal VOC 2012 dataset are shown in \cref{tab:voc_increment}. The performance of our GSC model outperforms other competing methods by 2.6\% mIoU in the 10-1 \textit{disjoint} scenario, which shows that our proposed GSC model is robust on long-sequence tasks.

\textbf{ADE20K:} To further verify the robustness of our GSC model, we conduct experiments on ADE20K dataset ~\cite{adedataset}. Under the \textit{overlapped} setting, \cref{tab:ade} shows quantitative results on 100-50, 100-10 and 50-50 scenarios. Our proposed GSC model outperforms previous methods in all experiments and achieves 0.5\% improvement on the challenging 50-50 scenario. This is due to the effectiveness of the proposed step-aware gradient compensation, soft-sharp semantic relation distillation and prototypical pseudo re-labeling strategies.

\begin{table}[htbp]
\caption{The mIoU(\%) of the last step on the Cityscapes dataset for incremental domain segmentation. The\textbf{\color{purple}red} denotes the highest results and the \textbf{\color{blue}blue} denotes the second highest results.}
\footnotesize
\setlength\tabcolsep{6.5pt}
\centering
\scalebox{1}{
\begin{tabular}{c|c|c|c}
\toprule
\textbf{Method} & \textbf{11-5 (3 steps)} & \textbf{11-1 (11 steps)} & \textbf{1-1 (21 steps)} \\ \midrule
FT              & 61.7                    & 60.4                     & 42.9                    \\ 
LWF~\cite{LWF}             & 59.7                    & 57.3                     & 33.0                    \\ 
LWF-MC~\cite{icarl}          & 58.7                    & 57.0                     & 31.4                    \\ 
ILT~\cite{ILT}             & 59.1                    & 57.8                     & 30.1                    \\ 
MiB~\cite{MIB}             & 61.5                    & 60.0                     & 42.2                    \\ 
PLOP~\cite{PLOP}            & 63.5                    & 62.1                     & 45.2                    \\ 
RCIL~\cite{RCIL}            & \textbf{\color{blue}64.3}                    & \textbf{\color{blue}63.0}                     & \textbf{\color{blue}48.9}                    \\ \midrule
\textbf{GSC} (ours)   & \textbf{\color{purple}65.7}                        & \textbf{\color{purple}63.6}                         & \textbf{\color{purple}55.8}                         \\ 
\bottomrule
\end{tabular}
}
\label{tab:cityscapes}
\end{table}

\subsection{Incremental Domain Segmentation}
Incremental domain segmentation is also an essential direction in incremental learning. As shown in \cref{tab:cityscapes}, our method outperforms the state-of-the-art methods on all experimental scenarios on the Cityscapes dataset ~\cite{cityscapesdataset}. Specifically, our method outperforms the state-of-the-art methods by 6.9\% mIoU in the 1-1 scenario. For this scenario, we display the performance of each step in \cref{fig:he} (right). However, several methods (\emph{e.g.,} ILT~\cite{ILT} and MiB~\cite{MIB}) that are effective under the setting of incremental class segmentation perform even worse than FT under the setting of incremental domain segmentation. This is because MiB only mitigates the background shift problem that does not exist in incremental domain segmentation, and ILT pays too much attention to restricting the forgetting of old knowledge and ignores the learning of new knowledge. By contrast, the proposed GSC model can achieve outstanding performance even in domain incremental semantic segmentation. 

\begin{table}[htbp]
\caption{Ablation study w.r.t. step-aware gradient compensation (SG), soft semantic relation distillation (SR), sharp confidence (SC), and prototypical pseudo re-labelling (PR) in the 10-1 disjoint scenario.}
\centering
\setlength\tabcolsep{9pt}
\scalebox{0.915}{
\begin{tabular}{ccccc|cc|c}
\toprule
\multicolumn{5}{c|}{\textbf{Method}}

& \multicolumn{3}{c}{\textbf{10-1 Disjoint}}\\ \midrule
PLOP~\cite{PLOP} & PR & SG & SR  & SC & 0-10          & 11-20           & all         \\ \midrule
 \checkmark                 &                              &                                &     &                           & \, 9.7          & \, 7.0         & \, 8.4     \\ 
 \checkmark & & &\checkmark & & 20.4  & \, 6.4  & 13.8    \\ 
 \checkmark & & & &\checkmark &20.4 & \, 5.9 & 13.5   \\ 
 \checkmark                 &                              &                                &\checkmark   &\checkmark                             & 21.3          & \, 9.6         & 15.8   \\ 
 \checkmark                 & \checkmark                             &                                &        &                        & 21.8          & \, 7.8         & 15.1   \\ 
 \checkmark &\checkmark & &\checkmark & &21.8 & \, 9.7 & 16.0   \\ 
 \checkmark &\checkmark & & &\checkmark & 23.0  & \, 9.1  & 16.4   \\
 \checkmark                 &\checkmark                              &                                &\checkmark    &\checkmark                            & 23.5          & \, 9.9         & 17.0 \\
 \checkmark                 &\checkmark                              &\checkmark                                &           &                     & 25.1          & \, 8.4         & 17.1   \\
 \checkmark &\checkmark &\checkmark &\checkmark & &29.2 & \, 8.2 & 19.2   \\
\checkmark &\checkmark &\checkmark & &\checkmark &26.7 & \, 9.0 & 18.3  \\
\checkmark                 & \checkmark                             & \checkmark                               & \checkmark             &\checkmark                  & 34.2          & \, 6.1        & 20.8 \\ \bottomrule
\end{tabular}}

\label{tab:ablation_study}
\end{table}

\subsection{Ablation Study}
To demonstrate the effectiveness of the proposed step-aware gradient compensation (SG), soft semantic relation distillation (SR), sharp confidence (SC) and prototypical pseudo re-labeling (PR), we conduct ablation studies in the 10-1 \textit{disjoint} scenario on the Pascal VOC 2012 dataset. Extensive results are shown in \cref{tab:ablation_study}, we observe that the mIoU is improved by 6.7\% via using PR to replace the pseudo-labeling strategy of the PLOP~\cite{PLOP} baseline. This shows that PR can produce high-quality pseudo labels for old classes by effectively removing misclassified pixels, which eases the problem of background shift. When performing ablation studies on SG, we use PR to generate pseudo labels as the basis for calculating SG. SG achieves 2.0\% mIoU improvement, due to overcoming forgetting by re-weighting gradient back-propagation of old classes. Meanwhile, removing SR and SC from our proposed GSC model decreases 2.5\% and 1.6\% mIoU performance by exploring the underlying semantic similarity relationship between classes. Overall, each module in our model is effective, and using the above modules simultaneously helps our model achieve the best performance. 

\subsection{Robustness to Class Order}

 As we all know, incremental learning methods are susceptible to the class order. In order to verify the robustness of our GSC model compared with other methods in terms of different class orders, we experiment with five different class orders of the 15-1 \textit{disjoint} scenario on the Pascal VOC 2012~\cite{RCIL} dataset. The class orders are provided as follows:

\noindent \textbf{A}: \{[0; 1; 2; 3; 4; 5; 6; 7; 8; 9; 10; 11; 12; 13; 14; 15]; [16]; [17]; [18]; [19]; [20]\};

\noindent \textbf{B}: \{[0; 12; 9; 20; 7; 15; 8; 14; 16; 5; 19; 4; 1; 13; 2; 11]; [17]; [3]; [6]; [18]; [10]\};

\noindent \textbf{C}: \{[0; 13; 19; 15; 17; 9; 8; 5; 20; 4; 3; 10; 11; 18; 16; 7]; [12]; [14]; [6]; [1]; [2]\};

\noindent \textbf{D}: \{[0; 15; 3; 2; 12; 14; 18; 20; 16; 11; 1; 19; 8; 10; 7; 17]; [6]; [5]; [13]; [9]; [4]\};

\noindent \textbf{E}: \{[0; 7; 5; 3; 9; 13; 12; 14; 19; 10; 2; 1; 4; 16; 8; 17]; [15]; [18]; [6]; [11]; [20]\}.

Experimental results are shown in \cref{tab:class_order}. Our proposed GSC model is more robust to class orders than state-of-the-art methods and obtains the best performance.

\begin{table}[htbp]
\caption{The mIoU(\%) of the last step for the 15-1 disjoint scenario on the Pascal VOC 2012 dataset. \textit{Average}: the mean mIoU(\%) and standard variance over five different class orders. \textbf{\color{purple}Red}: the highest results. \textbf{\color{blue}Blue}: the second highest results. $*$ represents results coming from re-implementation.}
\setlength\tabcolsep{7.2pt}
\centering
\small
\scalebox{0.915}{
\begin{tabular}{cccccc|c}
\toprule
& \multicolumn{6}{c}{\textbf{15-1 Disjoint}}      \\ \cmidrule{2-7}
\textbf{Method} & {A}     & {B}     & {C}     & {D}     & {E}     & {Average} \\ \midrule
\multicolumn{1}{c|}{ILT*~\cite{ILT}}             & \, 7.9  & 20.7 & \, 6.4  & 10.9 & 13.8 & 11.9 $\pm$ 5.1   \\ 
\multicolumn{1}{c|}{MiB*~\cite{MIB}}             & 39.9 & 23.7 & 34.3 & 40.6 & 48.0 & 37.3 $\pm$ 8.1   \\ 
\multicolumn{1}{c|}{PLOP*~\cite{PLOP}}           & 46.5 & 41.7 & 48.0 & 46.8 & 37.9 & 44.2 $\pm$ 3.8  \\ 
\multicolumn{1}{c|}{RCIL*~\cite{RCIL}}           & 51.7 & 51.1 &48.7  &56.2  &53.8  &\textbf{\color{blue}52.3 $\pm $ 2.5}   \\ \midrule
\multicolumn{1}{c|}{\textbf{GSC} (ours)}        &55.8       & 50.8      &51.9       & 55.2       &50.5       &\textbf{\color{purple}52.8 $\pm$ 2.2}      \\ \bottomrule
\end{tabular}}
\label{tab:class_order}
\end{table}

\subsection{Visualization on Pseudo Labels} 
\begin{figure}[t]
\centering
\hspace{-3mm}
\subfigure[Image]{
\label{fig:label:a}
\includegraphics[scale=0.665]{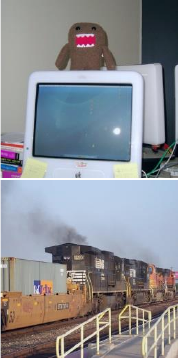}}
\hspace{-2.6mm}
\subfigure[PL~\cite{PLOP}]{
\label{fig:label:b}
\includegraphics[scale=0.665]{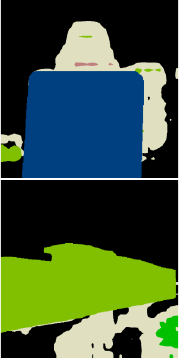}} 
\hspace{-2.6mm}
\subfigure[GSC (Ours)]{
\label{fig:label:c}
\includegraphics[scale=0.665]{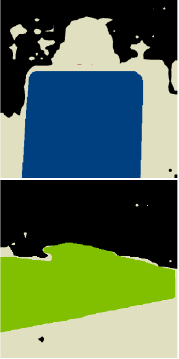}}
\hspace{-2.6mm}
\subfigure[GT]{
\label{fig:label:d}
\includegraphics[scale=0.665]{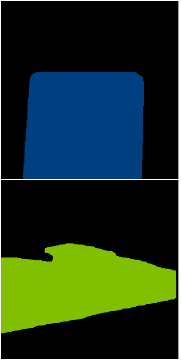}}
\caption{The visualization results of pseudo labels between our prototypical pseudo re-labeling and pseudo-labeling strategy ~\cite{PLOP} (PL) for 15-1 overlapped senario on Pascal VOC 2012 dataset. \textcolor{border}{White} represents ignored pixels.} 
\label{fig:pseudo_labels}
\end{figure}

\cref{fig:pseudo_labels} visualizes the results for pseudo labels of our prototypical pseudo re-labeling approach, compared with the pseudo-labeling strategy~\cite{PLOP} (PL). As shown in \cref{fig:pseudo_labels}, PL misclassifies the pixels of \texttt{background} to {\texttt{train}} (in the top row), {\texttt{sofa}} (in the bottom row) and {\texttt{person}} (in the top row), respectively. However, our prototypical pseudo re-labeling strategy can effectively remove misclassified pixels and regard them as ignored pixels, since it can rectify the pseudo labels by measuring the relative feature distances between pixels and all class-aware prototypes.

\subsection{Effectiveness of Step-aware Gradient Compensation}

\begin{figure}[htbp]
  \centering
  \includegraphics[width=1\linewidth]{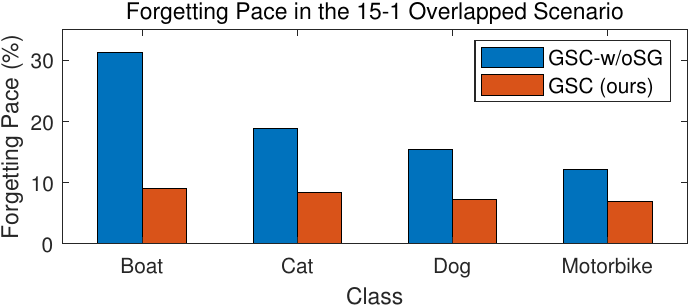}
   \caption{The Forgetting pace w.r.t. step-aware gradient compensation. Forgetting pace is the discrepancy between the class IoU after the first step and the last step. GSC-w/oSG is GSC without step-aware gradient compensation.}
   \label{fig:forgetting}
\end{figure}

 To verify the effectiveness of our step-aware gradient compensation which aims to make forgetting paces balanced. We compare our GSC and GSC-w/o SG (\emph{i.e.,}  GSC without step-aware gradient compensation) in \cref{fig:forgetting}. The results show that our step-aware gradient compensation effectively balances forgetting paces from 12.1\%$\sim$31.3\% mIoU to 6.9\%$\sim$9.0\% mIoU. This improvement validates the effectiveness of our step-aware gradient compensation due to re-weighting gradient back-propagation of different classes. 

\subsection{Hyperparameters Tuning}
We introduce two hyperparameters $\lambda_1$ and $\lambda_2$ in this paper. \cref{tab:4} shows results with more combinations of $\lambda_1$ and $\lambda_2$ in the 10-10 \textit{overlapped} scenario on the Pascal VOC 2012 dataset, but only negligible performance changes.
The mIoU by $\lambda_1=1$ and $\lambda_2=0.1$ is 70.4\%, even higher than 70.3\% in the paper. This demonstrates that our method is less affected by the hyperparameters $\lambda_1$ and $\lambda_2$ we introduce. Except $\lambda_1$ and $\lambda_2$, other hyperparameters (\emph{e.g.,} $\lambda$) and settings (\emph{e.g.,} validation dataset) follow other state-of-the-art methods (\emph{e.g.,} PLOP~\cite{PLOP} and RCIL~\cite{RCIL}).

\begin{table}[htbp]
\caption{Results for different combinations of $\lambda_1$ and $\lambda_2$.}

\setlength\tabcolsep{9pt}
\centering
\small
\scalebox{0.915}{
\begin{tabular}{cccccc}
\toprule

\multirow{2}{*}{{\textbf{Hyperparameter $\lambda_2$}}}&  \multicolumn{5}{c}{\textbf{Hyperparameter $\lambda_1$}}  \\ \cmidrule{2-6}
           & 1.0         & 0.8            & 0.5         & 0.3    & 0.1 \\ \midrule

\multicolumn{1}{c|}{1.0}        & 70.3        & 70.2          & 70.3        & 70.2    & 70.2 \\ \midrule
\multicolumn{1}{c|}{0.1}       & 70.4        & 70.4          & 70.2        & 70.3     &70.3\\ 
\bottomrule
\end{tabular}}

\label{tab:4}
\end{table}

\subsection{Combination with Other Methods}

It's worth noting that our GSC is plug-and-play. By applying our GSC to RCIL~\cite{RCIL}, \cref{tab:2} shows 6.2\% and 4.9\% mIoU gain compared to RCIL~\cite{RCIL} in the 10-1 \textit{disjoint} and 10-1 \textit{overlapped} scenarios, respectively. This demonstrates that our GSC has a large impact when combining with other methods.

\begin{table}[htbp]
\caption{Results by applying GSC to other methods. \textbf{\color{purple}Red} denotes the highest results.}

\setlength\tabcolsep{6pt}
\centering
\small
\scalebox{0.915}{
\begin{tabular}{ccc|c|cc|c}
\toprule
\multirow{2}{*}{{\textbf{Method}}}&  \multicolumn{3}{c|}{\textbf{10-1 Disjoint}} & \multicolumn{3}{c}{\textbf{10-1 Overlapped}}  \\ \cmidrule{2-7}
& 0-10        & 11-20       & all       & 0-10        & 11-20        & all \\ \midrule
\multicolumn{1}{c|}{RCIL} & 30.6        & \ \ 4.7        & 18.2      & 55.4        &  15.1        &  34.3    \\ \midrule
\multicolumn{1}{c|}{RCIL+\textbf{GSC} (ours)}     & 39.4        & \ \ 7.9       & \textbf{\color{purple}24.4}      & 55.4       & 21.5         & \textbf{\color{purple}39.2}    \\ 
\bottomrule
\end{tabular}}

\label{tab:2}
\end{table}

\section{Conclusion}
\label{sec:conclusion}
This paper proposes a \textbf{\underline{G}}radient-\textbf{\underline{S}}emantic \textbf{\underline{C}}ompensation (\textbf{GSC}) model to mitigate catastrophic forgetting and background shift for ISS from both gradient and semantic perspectives. Specifically, a step-aware gradient compensation based on pseudo labels is designed to balance the forgetting paces of old classes caused by imbalanced gradient back-propagation. Then we propose a soft semantic relation distillation loss and a sharp confidence loss to explore the semantic relations between old and new classes, further alleviating catastrophic forgetting. Furthermore, the prototypical pseudo re-labeling is developed to label old classes in the background for tackling background shift. Extensive experiments on three public segmentation datasets, \emph{i.e.,} Pascal VOC 2012, ADE20K, and Cityscapes, demonstrate the effectiveness of our GSC model.

From the aspect of algorithm design, the step-aware weight for the new step is set to $1$ simply in our current step-aware gradient compensation. We believe that the model can achieve better performance by exploring learnable step-aware weight for the new step in the future. From the aspect of applications, we currently verify the effectiveness and generalization of our proposed GSC model on several publicly available datasets. However, incremental semantic segmentation has broad prospects for medical imaging ~\cite{medical} and autonomous driving ~\cite{driving}. Therefore, we will try to apply the proposed GSC model to the above fields in the future.


%

\ifCLASSOPTIONcaptionsoff
  \newpage
\fi



\bibliographystyle{IEEEtran}

\bibliography{egbib}
%

%

\end{document}